# Designing Competent Mutation Operators via Probabilistic Model Building of Neighborhoods

**Kumara Sastry**
**David E. Goldberg**

# Designing Competent Mutation Operators via Probabilistic Model Building of Neighborhoods

Kumara Sastry[1,2], David E. Goldberg[1,3]
[1]Illinois Genetic Algorithms Laboratory (IlliGAL)
[2]Department of Material Science and Engineering
[3]Department of General Engineering
University of Illinois at Urbana-Champaign
Urbana, IL 61801
{ksastry,deg}@uiuc.edu

**Abstract**

This paper presents a *competent* selectomutative genetic algorithm (GA), that adapts linkage and solves hard problems quickly, reliably, and accurately. A probabilistic model building process is used to automatically identify key building blocks (BBs) of the search problem. The mutation operator uses the probabilistic model of linkage groups to find the best among competing building blocks. The competent selectomutative GA successfully solves additively separable problems of bounded difficulty, requiring only subquadratic number of function evaluations. The results show that for additively separable problems the probabilistic model building BB-wise mutation scales as $\mathcal{O}(2^k m^{1.5})$, and requires $\mathcal{O}(\sqrt{k} \log m)$ less function evaluations than its selectorecombinative counterpart, confirming theoretical results reported elsewhere (Sastry & Goldberg, 2004).

## 1 Introduction

One of the key challenges in designing an effective mutation operator is ensuring that it searches in the correct neighborhood. Existing mutation operators usually search in the local neighborhood of an individual, without taking into account the *global* neighborhood information. Recently, it was shown that a selectomutative algorithm that performs hillclimbing in building-block space can successfully solve boundedly-difficult problems in polynomial time as opposed to exponential time of simple mutation operators (Sastry & Goldberg, 2004). The results also showed that for additively separable search problems with deterministic fitness functions, building-block-wise mutation provided significant speed-up over recombination. The analysis assumed that both mutation and crossover operators had linkage information.

While several *competent* recombination operators that adapt linkage have been successfully and systematically designed, little attention has been paid to the development of competent mutation operators. Similarly, in local-search literature, while the importance of using a good neighborhood operator is often highlighted (Barnes, Dimova, & Dokov, 2003; Watson, 2003), there are no systematic methods for designing neighborhood operators that can solve a broad class of bounding problems.

This paper systematically designs a *competent* mutation operator that adapts linkage and performs local search in the building-block space. The important substructures are automatically identified using probabilistic models developed via machine-learning techniques. Specifically, we use the probabilistic-model-building procedure of extended compact genetic algorithm (eCGA) (Harik, 1999) to identify the linkage groups (or *global* neighborhood) of a search problem.

This paper is organized as follows. The next section gives a brief literature review, followed by a discussion on the relation between neighborhood operators and linkage groups. We then introduce eCGA followed by a description of the BB-wise mutation algorithm and provide empirical results on the scalability of each algorithm. Finally, we outline future research directions followed by conclusions.

## 2 Literature Review

One of the key challenges in the area of genetic and evolutionary algorithms is the systematic design of genetic operators with demonstrated scalability. One such design-decomposition theory for developing effective GA designs has been proposed (Goldberg, 1991; Goldberg, Deb, & Clark, 1992; Goldberg, 2002). Based on the design-decomposition theory many competent GAs have been designed, which can be broadly classified into three categories:

**Perturbation techniques** include the messy GA (Goldberg, Korb, & Deb, 1989), fast messy GA (Goldberg, Deb, Kargupta, & Harik, 1993), gene expression messy GA (Kargupta, 1996), linkage identification by nonlinearity check GA, and linkage identification by monotonicity detection GA (Munetomo & Goldberg, 1999), and dependency structure matrix driven genetic algorithm (DSMDGA) (Yu, Goldberg, Yassine, & Chen, 2003).

**Linkage adaptation techniques** such as linkage learning GA (Harik & Goldberg, 1997; Chen & Goldberg, 2002

**Probabilistic model building techniques** (Pelikan, Lobo, & Goldberg, 2002; Pelikan, 2002; Larrañaga & Loz such as population-based incremental learning (Baluja, 1994), the bivariate marginal distribution algorithm (Pelikan & Mühlenbein, 1999), the extended compact GA (eCGA) (Harik, 1999), iterated distribution estimation algorithm (Bosman & Thierens, 1999), Bayesian optimization algorithm (BOA) (Pelikan, Goldberg, & Cantú-Paz, 2000b; Pelikan & Goldberg, 2001).

While many of the above techniques are selectorecombinative GAs, little attention has been paid to the systematic design of selectomutative GAs that utilize linkage (or neighborhood) information. Recently, the authors demonstrated that a mutation operator that performs local search in building-block neighborhood takes problems that were intractable by a fixed mutation operator and renders them *tractable* (Sastry & Goldberg, 2004), requiring only polynomial number of function evaluations.

## 3 Neighborhood Operators and Linkage Groups

In local search literature, researchers have often recognized the importance of a good neighborhood operator in determining the effectiveness of a search algorithm (Barnes, Dimova, & Dokov, 2003; Watson, 2003; Vaughan, Jacobson, & Armstrong, 2000; Armstrong & Jacobson, 2004; Glover & Laguna, 1997).

A neighborhood operator that is capable of not only efficiently sampling the local neighborhood, but also able to jump to new less sampled neighborhoods, has often yielded good results (Armstrong & Jacobson, 2004; Glover & Laguna, 1997). Oftentimes the neighborhood operators are designed for a particular search problem on an ad-hoc basis. There are no systematic procedures or analytical guidance for designing a good neighborhood operator that can work on a broad class of search problems.

In genetic algorithms, while significant attention is paid to the design of recombination operators, little or no attention is paid to the design of mutation operators. In GAs, mutation is usually a secondary search operator which performs random walk *locally* around a solution. On the other hand, in evolution strategies (ES) (Rechenberg, 1973), in which mutation is the primary search operator, significant attention has been paid to the development of mutation operators. Several mutation operators, including adaptive techniques, have been proposed (Rechenberg, 1973; Schwefel, 1977; Bäck, 1996; Beyer, 1996; Hansen & Ostermeier, 2001). While the mutation operators used in ES are powerful search operators, the neighborhood information is still *local* around a single or few solutions.

However, for solving boundedly difficult GA-hard problems, local neighborhood information is not sufficient, and a mutation operator which uses local neighborhoods requires exponential time (Mühlenbein, 1992). Therefore, we utilize *machine-learning* tools and a population of candidate solutions of the search problem for automatically building *global* neighborhood (or linkage) information into the mutation operator. Unlike, adaptive mutation techniques in ES, which usually have local neighborhood information adapted over time, our method leads to a more global induction of the neighborhood. Specifically, we build probabilistic models of *global* neighborhood information by sampling candidate solutions to the search problem. The mutation operator proposed in this paper utilizes the *global* neighborhood information to search among competing sub-solutions.

The procedure used to build the neighborhood information is based on the model-building procedure of eCGA, which is explained in the following section.

## 4 Extended Compact Genetic Algorithm (eCGA)

The extended compact GA proposed by Harik (Harik, 1999) is based on a key idea that the choice of a good probability distribution is equivalent to linkage learning. The measure of a good distribution is quantified based on minimum description length (MDL) models. The key concept behind MDL models is that all things being equal, simpler distributions are better than more complex ones. The MDL restriction penalizes both inaccurate and complex models, thereby leading to an optimal probability distribution. Thus, MDL restriction reformulates the problem of finding a good distribution as an optimization problem that minimizes both the probability model as well as population representation. The probability distribution used in eCGA is a class of probability models known as marginal product models (MPMs). MPMs are formed as a product of marginal distributions on a partition of the genes. MPMs also facilitate a direct linkage map with each partition separating tightly linked genes. For example, the following MPM, [1,3][2][4], for a four-bit problem represents that the $1^{st}$ and $3^{rd}$ genes are linked and $2^{nd}$ and $4^{th}$ genes are independent.

The eCGA can be algorithmically outlined as follows:

1. Initialization: The population is usually initialized with random individuals.

2. Evaluate the fitness value of the individuals

3. Selection: The eCGA uses an s-wise tournament selection (Goldberg, Korb, & Deb, 1989).

4. Build the probabilistic model: In eCGA, both the structure and the parameters of the model are searched. A greedy search heuristic is used to find an optimal model of the selected individuals in the population.

5. Create new individuals: In eCGA, new individuals are created by sampling the probabilistic model.

6. Replace the parental population with the offspring population.

7. Repeat steps 2–6 until some convergence criteria are met.

Two things need further explanation, one is the identification of MPM using MDL and the other is the creation of a new population based on MPM. The identification of MPM in every generation is formulated as a constrained optimization problem, that minimizes the sum of the model complexity, $C_m$, which represents the cost of a complex model and compressed population complexity, $C_p$, which represents the cost of using a simple model as against a complex one.

In essence, the model complexity, $C_m$, quantifies the model representation size in terms of number of bits required to store all the marginal probabilities. Let, a given problem of size $\ell$ with binary alphabets, have $m$ partitions with $k_i$ genes in the $i^{\text{th}}$ partition, such that $\sum_{i=1}^{m} k_i = \ell$. Then each partition $i$ requires $2^k - 1$ independent frequencies to completely define its marginal distribution. Furthermore, each frequency is of size $\log_2(n)$, where $n$ is the population size. Therefore, the model complexity (or the representation size), $C_m$, is given by

$$C_m = \log_2(n) \sum_{i=1}^{m} \left(2^{k_i} - 1\right). \tag{1}$$

The compressed population complexity, $C_p$, quantifies the data compression in terms of the entropy of the marginal distribution over all partitions. Therefore, $C_p$ is evaluated as

$$C_p = n \sum_{i=1}^{m} \sum_{j=1}^{2^{k_i}} -p_{ij} \log_2(p_{ij}) \tag{2}$$

where $p_{ij}$ is the frequency of the $j^{\text{th}}$ gene sequence of the genes belonging to the $i^{\text{th}}$ partition. In other words, $p_{ij} = N_{ij}/n$, where $N_{ij}$ is the number of chromosomes in the population (after selection) possessing bit-sequence $j \in [1, 2^{k_i}]$ [1] for $i^{\text{th}}$ partition.

The following greedy search heuristic is used to find an optimal or near-optimal probabilistic model:

1. Assume each variable is independent of each other.

2. Compute the model complexity and population complexity values of the current model.

3. Consider all possible $\frac{1}{2}\ell(\ell - 1)$ merges of two variables.

---
[1] Note that a BB of length $k$ has $2^k$ possible sequences where the first sequence denotes be $00\cdots0$ and the last sequence $11\cdots1$

4. Evaluate the model and compressed population complexity values for each model structure.

5. Select the merged model with lowest combined complexity.

6. If the combined complexity of the best merged model is better than the combined complexity of the model evaluated in step 2., replace it with the best merged model and go to step 2.

7. If the combined complexity of the best merged model is less than or equal to the combined complexity, the model cannot be improved and the model of step 2. is the probabilistic model of the current generation.

The offspring population are generated by randomly choosing subsets from the current individuals according to the probabilities of the subsets as calculated in the probabilistic model.

Analytical models have been developed for predicting the population-sizing and the scalability of probabilistic model building GAs (Pelikan, Goldberg, & Cantú-Paz, 2000a; Pelikan, Sastry, & Goldberg, 2003). The models predict that the population size required to solve a problem with $m$ building blocks of size $k$ with a failure rate of $\alpha = 1/m$ is given by

$$n \propto 2^k \left(\frac{\sigma_{BB}}{d}\right) m \log m, \tag{3}$$

and the number of function evaluations is given by

$$n_{fe} \propto \left(\frac{\sigma_{BB}}{d}\right) \sqrt{k} \cdot 2^k m^{1.5} \log m, \tag{4}$$

where $\sigma_{BB}$ is fitness-variance of a BB and $d$ is the signal difference between competing BBs (Goldberg, Deb, & Clark, 1992).

Equations 3 and 4 are verified with empirical results for the m k-deceptive function (Goldberg, 1987; Deb & Goldberg, 1993) with *loose linkage* in Figures 1(a) and 1(b). By loose linkage we mean that the components of a BB are located far apart from each other in the chromosome. Fixed recombination operators such as one-point crossover or uniform crossover need exponential time to solve such loosely-linked deceptive problems (Thierens & Goldberg, 1993). The results show that eCGA automatically identifies the linkage groups and solve additively separable GA-hard problems, requiring only subquadratic number of function evaluations.

In obtaining the empirical results, we use a tournament selection with tournament size of 8. An eCGA run is terminated when all the individuals in the population converge to the same fitness value. The average number of BBs correctly converged are computed over 30 independent runs. The minimum population size required such that $m-1$ BBs converge to the correct value is determined by a bisection method (Sastry, 2001). The results of population-size is averaged over 30 such bisection runs, while the results for the function-evaluation ratio is averaged over 900 independent eCGA runs.

## 5  Probabilistic Model Building BB-wise Mutation

As explained in the previous section, eCGA builds marginal product models that yields a direct mapping of linkage groups among successful individuals. Therefore, for BB identification purposes, we use the model-building procedure of eCGA. Once the linkage-groups are identified, we use an

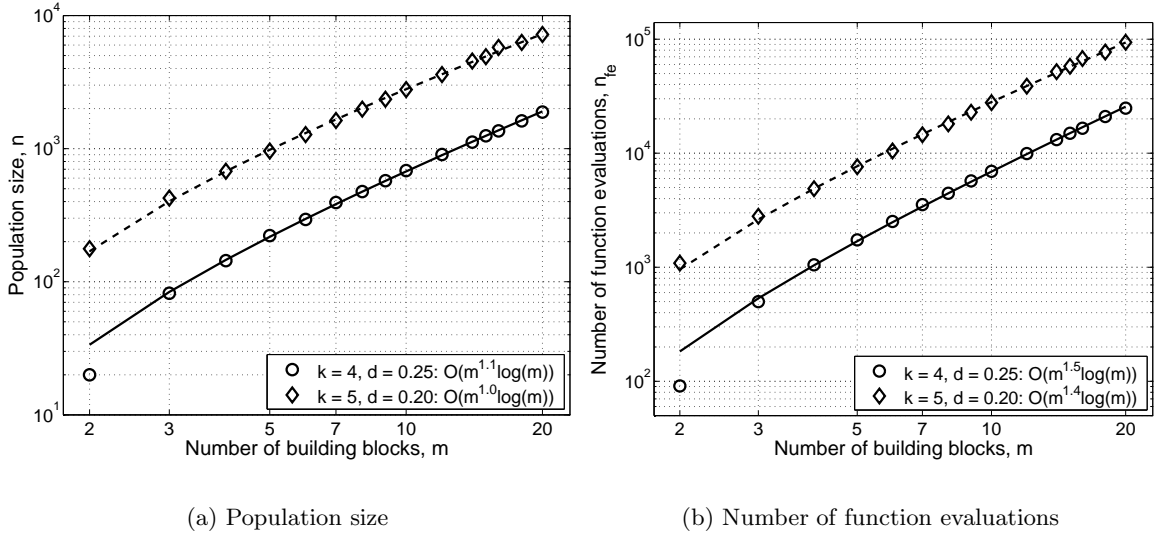

(a) Population size

(b) Number of function evaluations

Figure 1: Population size (Equation 3) and number of function evaluations (Equation 4) required by eCGA to successfully solve m k-Trap function. The results are averaged over 900 eCGA runs for the function evaluations and 30 bisection runs for the population size. The relative deviation for the empirical results is less than 1%. The results show that the population size scales as $\mathcal{O}(2^k m \log m)$ and the number of function evaluations scales as $\mathcal{O}(2^k m^{1.5} \log m)$.

*enumerative BB-wise mutation* operator as used elsewhere (Sastry & Goldberg, 2004). For example, if model builder identifies $m$ BBs of size $k$ each, the BB-wise algorithm will select the best BB out of $2^k$ possible ones in each of the $m$ partition. The detailed procedure of the competent selectomutative GA is given in the following:

1. Initialize the population with random individuals and evaluate their fitness.

2. Selection: This procedure is similar to that of eCGA as described in Section 4.

3. Build the probabilistic model as explained in Section 4 to obtain linkage-group information.

4. Use the best individual from the population for BB-wise mutation.

5. Consider the first non-mutated BB. Here the BB order is chosen arbitrarily from left-to-right, however, different schemes can be—or may required to be—chosen to decide the order of BBs. For example, BB partitions that contain most *active* variables might be mutated before those that contain less active variables.

6. Create $2^k - 1$ unique individuals with all possible schemata in the chosen BB partition. Note that the schemata in other partitions are the same as the original individual (from step 2).

7. Evaluate all $2^k - 1$ individuals and retain the best for mutation of BBs in other partitions.

8. Repeat steps 5–7 till BBs of all the partitions have been mutated.

Steps 1–3 are identical to the ones used in eCGA (Section 4) and steps 4–8 are similar to the BB-wise mutation operator used elsewhere (Sastry & Goldberg, 2004).

Note that the performance of the BB-wise mutation can be slightly improved by using a greedy heuristic to search for the best among competing BBs, however, as shown later, the scalability of the probabilistic model building BB-wise mutation operator is determined by the population-size required to accurately identify the building blocks.

It should also be noted that while eCGA can only build linkage groups with non-overlapping variables, the mutation procedure can be easily used with other linkage identification techniques that can handle overlapping BBs such as BOA (Pelikan, Goldberg, & Cantú-Paz, 2000b) or DSMDGA (Yu, Goldberg, Yassine, & Chen, 2003). However, since the effect of overlapping interactions between variables is similar to that of an exogenous noise (Goldberg, 2002), crossover is likely to be more effective than mutation (Sastry & Goldberg, 2004).

Finally, we perform linkage identification only once in the initial generation. This offline linkage identification works well on problems with BBs of nearly equal salience. However, for problems with BBs of non-uniform salience, we would have to perform linkage identification and update BB information in regular intervals. Furthermore, it might be more efficient to utilize both BB-wise mutation and eCGA model sampling simultaneously or sequentially along the lines of *hybridization* (Goldberg & Voessner, 1999; Sinha & Goldberg, 2003; Sinha, 2003) and *time-continuation* (Goldberg, 1999; Srivastava, 2002) techniques.

However, the objective of this paper is to couple linkage identification with a mutation operator that performs local search in the BB neighborhood and to verify its effectiveness in solving boundedly difficult additively separable problems. Moreover, the aforementioned enhancements can be designed on the proposed competent selectomutative GA.

## 5.1 Scalability of the BB-wise Mutation

The scalability of the selectomutative GA depends on two factors: (1) The population size required to build accurate probabilistic models of the linkage groups, and (2) the total number of evaluations performed by the BB-wise mutation operator to find optimum BBs in all the partitions.

Pelikan and Sastry (Pelikan, Sastry, & Goldberg, 2003) developed facetwise models for predicting the critical and maximum population-size required to correctly identify good interactions among variables. They showed that the minimum population size scales as

$$n_{\min} = \mathcal{O}\left(2^k m^{1.05}\right), \tag{5}$$

and the maximum population size which avoids discovery of false dependencies between independent variables is given by

$$n_{\max} = \mathcal{O}\left(2^k m^{2.1}\right), \tag{6}$$

In other words, to avoid incorrect identification of BBs, the population size should be less than $n_{\max}$. Since we require that *all* the BBs be correctly identified in the first generation itself, the population size required should be greater than $n_{\min}$, but less than $n_{\max}$. That is,

$$\mathcal{O}\left(2^k m^{1.05}\right) \leq n \leq \mathcal{O}\left(2^k m^{2.1}\right). \tag{7}$$

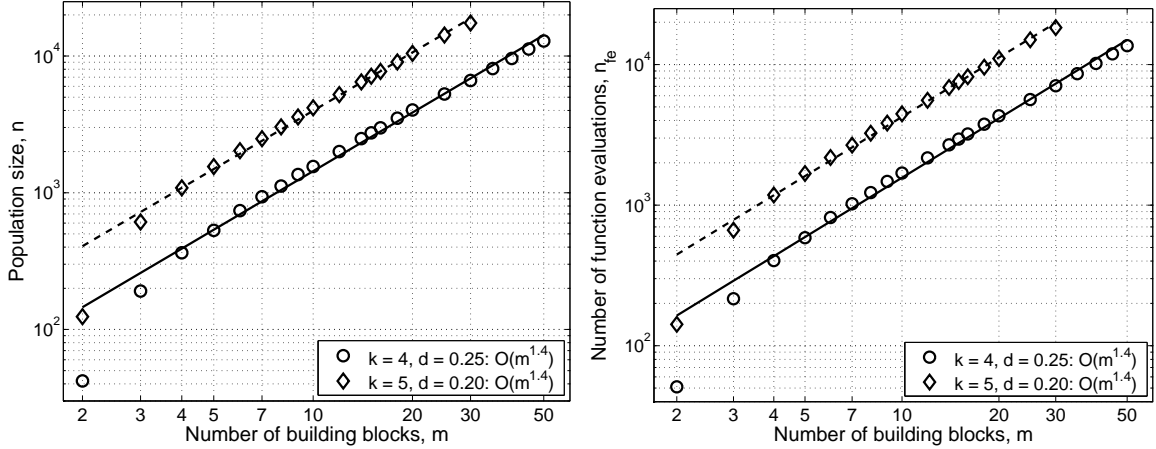

(a) Population size       (b) Number of function evaluations

Figure 2: Population size (Equation 7 and the number of function evaluations (Equation 10) required by BB-wise mutation for solving m k-Trap function. The results are averaged over 900 runs for the function evaluations and 30 bisection runs for the population size. The relative deviation for the empirical results is less than 0.2%. The population size and the number of function evaluations both scale as $\mathcal{O}(2^k m^{1.5})$.

Since the model building is performed only once, the total number of function evaluations scales as the population size. That is,

$$\mathcal{O}\left(2^k m^{1.05}\right) \leq n_{\text{fe},1} \leq \mathcal{O}\left(2^k m^{2.1}\right). \tag{8}$$

During BB-wise mutation, we evaluate $2^k - 1$ individuals for determining the best BBs in each of the $m$ partitions. Therefore, the total number of function evaluations used during BB-wise mutation is

$$n_{\text{fe},2} = \left(2^k - 1\right) m = \mathcal{O}\left(2^k m\right). \tag{9}$$

From Equations 8 and 9, the total number of function evaluations scales as

$$\mathcal{O}\left(2^k m^{1.05}\right) \leq n_{\text{fe}} \leq \mathcal{O}\left(2^k m^{2.1}\right). \tag{10}$$

We now empirically verify the scale-up of the population size and the number of function evaluations required for successfully solving the m k-trap problem with loose linkage in Figures 2(a) and 2(b), respectively. In contrast to fixed mutation operators which require $\mathcal{O}(m^k \log m)$ (exponential) number of function evaluations to solve additively separable GA-hard problems (Mühlenbein, 1992), the proposed eCGA-based BB-wise mutation operator that automatically identifies the linkage groups requires only $\mathcal{O}(2^k m^{1.5})$ (polynomial) number of evaluations.

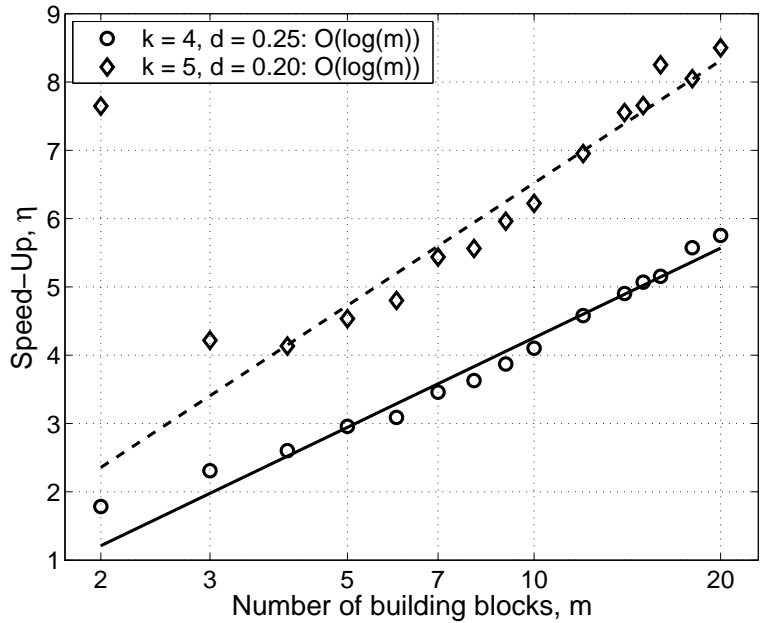

Figure 3: Empirical verification of the speed-up (Equation 11) obtained by using the probabilistic model building BB-wise mutation over eCGA for the m k-Trap function. The results show that the speed-up scales as $\mathcal{O}(\sqrt{k}\log m)$.

## 6 eCGA vs. Building-Block-wise Mutation

The previous two sections demonstrated the scalability of eCGA and the competent selectomutative GA. In this section we analyze the relative computational costs of using eCGA or the BB-wise mutation algorithm for successfully solving additively separable problems of bounded difficulty.

The results from the above sections (Equations 4 and 10) indicate that while the scalability of eCGA is $\mathcal{O}\left(2^k\sqrt{k}m^{1.5}\log m\right)$, the scalability of the BB-wise mutation is $\mathcal{O}\left(2^k m^{1.5}\right)$. Therefore, the probabilistic model building BB-wise mutation operator is $\mathcal{O}\left(\sqrt{k}\log m\right)$ faster than eCGA in solving boundedly difficult additively separable problems. That is, the speed-up—which is defined as the ratio of number of function evaluations required by eCGA to that required by the selectomutative GA—is given by

$$\eta = \frac{n_{\text{fe}}(\text{eCGA})}{n_{\text{fe}}(\text{BBwise Mutation})} = \mathcal{O}\left(\sqrt{k}\log m\right). \qquad (11)$$

Empirical results shown in Figure 3 agrees with the above equation. The results show that the probabilistic model building BB-wise mutation is $\mathcal{O}(\sqrt{k}m)$ times faster than the extended compact GA. The results are also in agreement with the analytical results derived for an *ideal* BB-wise operator (Sastry & Goldberg, 2004).

# 7   Future Work

We demonstrated the potential of inducting *global* neighborhood information into mutation operations via the automatic discovery of linkage groups by probabilistic model-building techniques. The results are very encouraging and warrants further research in one of more of the following avenues:

**Hybridization of competent crossover and mutation:** While we considered a bounding case of crossover vs. mutation, it is likely to be more efficient to use an efficient hybrid of competent crossover and mutation operators. For example, we can consider a hybrid GA with *oscillating* populations. A large population is used to gather linkage information and used for crossover, while a small population is used for searching in BB neighborhood.

**Problems with overlapping building blocks:** While this paper considered problems with non-overlapping building blocks, many problems have different building blocks that share common components. The performance of probabilistic model building BB-wise mutation on problems with overlapping building blocks have to be analyzed. Since the effect of overlapping variable interactions is similar to that of exogenous noise (Goldberg, 2002), based on our recent analysis (Sastry & Goldberg, 2004), a crossover is likely to be more useful than mutation.

**Problems with non-uniform BB salience:** In this paper we considered additively separable problems with uniform sub-solution salience. Unlike uniformly-scaled problems, in non-uniformly scaled problems BBs are identified sequentially over time. Therefore, in such cases, we would need to regularly update the BB information and develop theory to predict the updating schedule.

**Hierarchical problems:** One of the important class of nearly decomposable problems is hierarchical problems, in which the building-block interactions are present at more than a single level. Further investigation is necessary to analyze the performance of BB-wise mutation on hierarchical problems.

# 8   Summary & Conclusions

In this paper, we have introduced a systematic procedure for the automatic induction of *global* neighborhood information into the mutation operator via the discovery of linkage groups of a problem. We used probabilistic model building techniques to develop a probabilistic model of linkage information of a search problem. The BB-wise mutation operators uses the linkage (or neighborhood) information to perform local search among competing sub-solutions.

We derived an analytical bound and empirically verified the scalability of the competent mutation operator on boundedly-difficult additively separable problems. The results showed that the BB-wise mutation operator successfully solves GA-hard problems, requiring only subquadratic number of function evaluations. That is, for an additively separable problem with $m$ BBs of size $k$ each, the number of function evaluations scales as $\mathcal{O}(2^k m^{1.5})$. We also compared the probabilistic model-building mutation with probabilistic model-building crossover head to head. For deterministic additively separable problems, we showed that BB-wise mutation provides significant advantage over crossover. The results show that the speed-up of using BB-wise mutation over crossover on deterministic problems is $\mathcal{O}(\sqrt{k}\log m)$, which is in agreement with analytical results (Sastry & Goldberg, 2004).

## Acknowledgments

This work was sponsored by the Air Force Office of Scientific Research, Air Force Materiel Command, USAF, under grant F49620-00-0163 and F49620-03-1-0129, the National Science Foundation under grant DMI-9908252, ITR grant DMR-99-76550 (at Materials Computation Center), and ITR grant DMR-0121695 (at CPSD), and the Dept. of Energy through the Fredrick Seitz MRL (grant DEFG02-91ER45439) at UIUC. The U.S. Government is authorized to reproduce and distribute reprints for government purposes notwithstanding any copyright notation thereon.

The views and conclusions contained herein are those of the authors and should not be interpreted as necessarily representing the official policies or endorsements, either expressed or implied, of the Air Force Office of Scientific Research, the National Science Foundation, or the U.S. Government.

Pelikan, M., & Mühlenbein, H. (1999). The bivariate marginal distribution algorithm. In Roy, R., Furuhashi, T., & Chawdhry, P. K. (Eds.), *Advances in Soft Computing - Engineering Design and Manufacturing* (pp. 521–535). London: Springer-Verlag.

Pelikan, M., Sastry, K., & Goldberg, D. E. (2003). Scalability of the Bayesian optimization algorithm. *International Journal of Approximate Reasoning*, *31*(3), 221–258. (Also IlliGAL Report No. 2002024).

Rechenberg, I. (1973). *Evolutionsstrategie: Optimierung technischer systeme nach prinzipien der biologischen evolution*. Stuttgart: Frommann-Holzboog.

Sastry, K. (2001). *Evaluation-relaxation schemes for genetic and evolutionary algorithms*. Master's thesis, University of Illinois at Urbana-Champaign, General Engineering Department, Urbana, IL. (Also IlliGAL Report No. 2002004).

Sastry, K., & Goldberg, D. (2004, January). *Let's get ready to rumble: Crossover versus mutation head to head* (IlliGAL Report No. 2004005). Urbana, IL: University of Illinois at Urbana-Champaign.

Schwefel, H.-P. (1977). Numerische optimierung von computer-modellen mittels der evolutionsstrategie. *Interdisciplinary Systems Research*, *26*, .

Sinha, A. (2003). *Designing efficient genetic and evolutionary algorithm hybrids*. Master's thesis, University of Illinois at Urbana-Champaign, General Engineering Department, Urbana, IL. (Also IlliGAL Report No. 2003020).

Sinha, A., & Goldberg, D. E. (2003, January). *A survey of hybrid genetic and evolutionary algorithms* (IlliGAL Report No. 2003004). Urbana, IL: University of Illinois at Urbana-Champaign, General Engineering Department.

Srivastava, R. (2002). *Time continutation in genetic algorithms*. Master's thesis, University of Illinois at Urbana-Champaign, General Engineering Department, Urbana, IL. (Also IlliGAL Report No. 2001021).

Thierens, D., & Goldberg, D. E. (1993). Mixing in genetic algorithms. *Proceedings of the International Conference On Genetic Algorithms*, 38–45.

Vaughan, D., Jacobson, S., & Armstrong, D. (2000). A new neighborhood function for discrete manufacturing process design optimization using generalized hill climbing algorithms. *ASME Journal of Mechanical Design*, *122*(2), 164–171.

Watson, J.-P. (2003). *Empirical modeling and analysis of local search algorithms for the job-shop scheduling problem*. Doctoral dissertation, Colorado State University, Fort Collins, CO.

Yu, T.-L., Goldberg, D. E., Yassine, A., & Chen, Y.-P. (2003). A genetic algorithm design inspired by organizational theory: Pilot study of a dependency structure matrix driven genetic algorithm. *Artificial Neural Networks in Engineering*, 327–332. (Also IlliGAL Report No. 2003007).
13